# A non-extensive entropy feature and its application to texture classification

Seba Susan, Madasu Hanmandlu
Department of Electrical Engineering,
Indian Institute of Technology (IIT) Delhi, New Delhi, India

*Abstract:*
This paper proposes a new probabilistic non-extensive entropy feature for texture characterization, based on a Gaussian information measure. The highlights of the new entropy are that it is bounded by finite limits and that it is non additive in nature. The non additive property of the proposed entropy makes it useful for the representation of information content in the non-extensive systems containing some degree of regularity or correlation. The effectiveness of the proposed entropy in representing the correlated random variables is demonstrated by applying it for the texture classification problem since textures found in nature are random and at the same time contain some degree of correlation or regularity at some scale. The gray level co-occurrence probabilities (GLCP) are used for computing the entropy function. The experimental results indicate high degree of the classification accuracy. The performance of the new entropy function is found superior to other forms of entropy such as Shannon, Renyi, Tsallis and Pal and Pal entropies on comparison. Using the feature based polar interaction maps (FBIM) the proposed entropy is shown to be the best measure among the entropies compared for representing the correlated textures.

---

# 1. INTRODUCTION

Texture is an integral part of natural images and plays a significant role in pattern recognition, image processing and computer vision. Texture classification aims at assigning labels to the unknown textures represented by some textural features. The widely used approaches for the texture feature extraction include co-occurrence matrix [1,2], Markov random fields [3], Gabor filters [4,5] and Wavelet representations [6-11]. The recent works utilize features like wavelet packet neural network adaptive norm entropy [8] and log polar wavelet energy signatures [11] for the scale-invariant texture classification. In [27] Ves *et. al.* have proposed a structuring element for defining the granulometric size distribution of the texture from which descriptors like mean, variance, skewness and kurtosis are derived and used for texture classification. Strand and Taxt [12] have compared filtering techniques and co-occurrence statistics and found that co-occurrence matrix features were the best. Buf *et. al.* [13] report that many texture features offer roughly the same performance while evaluating co-occurrence matrices, fractal dimension, filter banks, gray level extrema per area and curvilinear integration features. Conners and Harlow [14] claim that the co-occurrence based features are better than run length difference, intensity differences and power spectra. Petrou *et. al.* [26] have also stated in their work that the co-occurrence based features perform best in terms of texture classification accuracy. Co-occurrence matrix finds the joint probability of occurrence of two pixel intensities separated by an offset distance. The resulting co-occurrence probabilities are used to calculate texture features. The use of entropy as a feature for characterizing texture is a well explored area of research. Various measures of entropy have been proposed in the past. Shannon [15] has defined the information gain from an event, as the negative of the logarithm of the probability of occurrence of the event. Shannon entropy is one of the fourteen co-occurrence features used by Haralick [1] for texture classification. Idrissa *et. al.* [2] retain eight of Haralick's co-occurrence features namely, variance, energy, standard deviation, skew, kurtosis, inverse difference moment, contrast and covariance, based on relative significance and computational cost. Renyi entropy [16] is an improvement over Shannon entropy.

Other forms of entropy were introduced for application to image processing. Tsallis [17] has generalized Shannon entropy for non-extensive physical processes by defining a pseudo-additive entropy for the statistically independent systems. Tsallis entropy is used successfully for image segmentation by thresholding in [18]. Pal and Pal [19] claim that an exponential entropy is more effective for the image specific applications than Shannon and Renyi logarithmic measures of entropy since the exponential entropic measure is bounded by finite upper and lower bounds unlike the logarithmic measure which is undefined when either the probability of occurrence of the event is zero or when the events are equally likely and the number of events is very large.

The probabilistic non-extensive entropy is proposed in this paper is based on a Gaussian information gain function. The properties of the proposed entropy are supported by proofs. The conditional, joint and relative entropies are defined as byproducts of the new entropy and their properties are investigated. The proposed entropy function is applied for texture identification and classification with successful results. Some related work regarding fuzzy entropy based on exponential functions has been done by the authors in [25], and this serves as a precursor to our main work. The organization of the paper is as follows: The formulation of the new entropy function is described in Section 2 together with its properties. Some definitions and properties springing forth from the new entropy function are explained in Section 3. Section 4 describes the application of non additive property of the proposed entropy for texture classification. Section 5 outlines the methodology for texture classification. Section 6 discusses the experimental results. Finally the conclusions from overall results are summarized in Section 7.

## 2. FORMULATING THE NEW ENTROPY FEATURE

### 2.1 Motivation for the proposed entropy function

In most of the cases the entropy function is the sum of products of linear probability terms and non-linear information gains. The non linearity of the information gain function plays an important role in the identification of textures having high spatial correlation and containing non additive information content. The exponential function is proved to be one of the best non-linear functions to be used as the information gain function [19]. In this work, we attempt to increase the non-linearity of the exponential information gain introduced in [19] by replacing the linear exponent of the exponential by a quadratic probability term. Alternatively, we can also justify the use of this Gaussian type information measure by the classical probability theory.

One of the axioms of classical probability theory states that in an experiment, as the probability of some event approaches 1 and the probabilities of all other events approach zero, the entropy or uncertainty regarding the experiment reduces to zero. It may be noted that since the information gain decreases monotonically to zero as the occurrence of an event becomes certain, it can be approximated by a one sided zero mean Gaussian function (since probability is always non-negative) with the standard deviation of $1/\sqrt{2}$. The non-linearity of the Gaussian information gain ensures that only the relevant information lying in the 'bell' of the curve will be considered for calculating the entropy and the rest will be discarded. The degree of relevance of the information is determined from the standard deviation of the Gaussian curve, which is set to $1/\sqrt{2}$ in order to simplify calculations. The Gaussian function by itself is well researched and its properties are well studied. This justifies our selection of the Gaussian function as the non-linear information gain function for the new non-extensive entropy proposed in this paper.

### 2.2 Definition of the new entropy function

Consider a random variable X={x1,x2,….,xn} with the probabilities P={p1,p2,….pn}. Assume that the probability distribution is complete i.e. $p_i \in [0,1]$ and $\sum_{i=1}^{n} p_i = 1$ for i=1,2,…n where, n is the number of probabilistic experiments. Let the information gain on the $^{ith}$ event of X with an associated probability $p_i$ be defined by the Gaussian function:

$$I(p_i) = e^{-p_i^2} \qquad (1)$$

The non-linear graph of information gain *I(p$_i$)* versus the probability *p$_i$* is shown in Fig. 1.

As can be seen from Fig. 1, a maximum information gain of 1 is achieved when an event is least likely to occur (*p$_i$ =0*). As the occurrence of an event becomes more certain (*p$_i$→1*), the information gain reduces monotonically to reach a minimum value of *e$^{-1}$*.

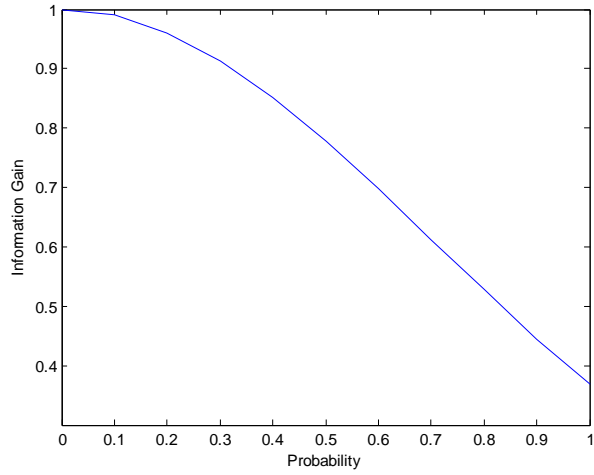

**Fig 1.**: *Plot of the Gaussian Information gain I(p$_i$) as the probability p$_i$ increases from 0 to 1.*

The entropy of X is defined as

$$H(P) = E(I(p_i)) = \sum_{i=1}^{n} p_i I(p_i) = \sum_{i=1}^{n} p_i e^{-p_i^2} \qquad (2)$$

The normalized entropy $H_N$ is of the form

$$H_N = \frac{(H - H_{\min})}{(H_{\max} - H_{\min})} \qquad (3)$$

i.e.
$$H_N = \frac{(H - e^{-1})}{e^{-\frac{1}{n^2}} - e^{-1}} \qquad (4)$$

The entropy in (4) is called the Non-Extensive entropy with a Gaussian Information gain.

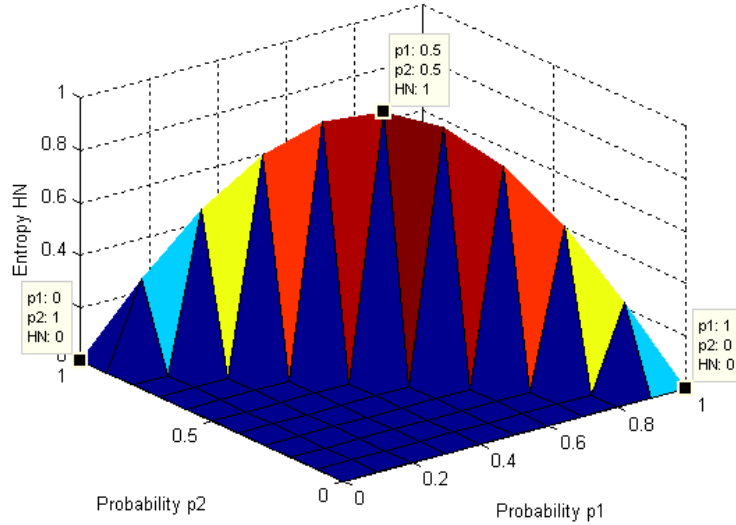

**Fig. 2**: *Normalized Entropy H$_N$ plotted as function of the probabilities p$_1$ and p$_2$ where p$_1$,p$_2$ε[0,1] and $\sum_{i=1}^{2} p_i = 1$.*

The values assigned to $H_{max}$ and $H_{min}$ in (3) are derived in the properties detailed below. The normalized entropy $H_N$ for a two state system is plotted in Fig. 2. It is observed from Fig.2 that the minimum values of $H_N$ occur when the probability of either event becomes one, while the maximum value is obtained when both the events are equally likely to occur.

## 2.3 Properties of the new entropy function

These are stated as follows:

*Property 1*: The Gaussian information measure $I(p_i) = e^{-p_i^2}$ is a continuous function for all $p_i \in [0,1]$

*Property 2*: $I(p_i)$ is bounded by the lower limit of $e^{-1}$ and the upper limit of 1.

*Property 3*: With the increase in the value of $p_i$, $I(p_i)$ decreases.

*Property 4*: The entropy $H(P) = \sum_{i=1}^{n} p_i I(p_i) = \sum_{i=1}^{n} p_i e^{-p_i^2}$ is a continuous function.

*Property 5*: Entropy is a concave function.

*Property 6*: Entropy is maximum when all $p_i$'s are equal, for i=1,2,…n. In other words, $H_{max} = H(\frac{1}{n}, \frac{1}{n}, \ldots, \frac{1}{n}) = e^{-\frac{1}{n^2}}$.

*Property 7*: Any move towards equalizing the probabilities increases the entropy. Consider a partition A = [$A_1$, $A_2$, …,$A_n$] with the probabilities, $p_i = Pr[A_i]$ and assume that $p_1 < p_2$. If $p_1$ is increased and $p_2$ is decreased by an equal amount, say $\delta$ ($\delta > 0$ and $\delta \leq \frac{p_2 - p_1}{2}$), then the entropy increases.

*Property 8*: Entropy is minimum if and only if all $p_i$'s except one are zeros and that single $p_i$ is equal to 1, i.e. $H_{min} = e^{-1}$

*Property 9*: If $p_1 = p_2 = \ldots = p_n = 1/n$ then $H(P)$ is an increasing function of n.

*Property 10*: Consider the partition of the event space as A = [$A_1$, $A_2$, …..,$A_n$] and the probability $p_i = Pr(A_i)$. If a new partition B is made by subdividing one of the sets of A, then $H(B) \geq H(A)$.

The proofs of all the above properties are given in Appendix.

## 3. SOME ADDITIONAL PROPERTIES OF PROPOSED ENTROPY FEATURE

### 3.1 Definitions of Conditional Entropy, Joint Entropy and Relative entropy

Consider two random variables X={$x_1,x_2,\ldots,x_n$} and Y={$y_1,y_2,\ldots y_n$} having finite complete probability vectors of {$p(x_1),p(x_2),\ldots,p(x_n)$} and {$p(y_1),p(y_2),\ldots,p(y_n)$} respectively. Then from (2) the proposed non-extensive entropy of X is given by:

$$H(X) = \sum_{x \varepsilon X} p(x_i) e^{-p^2(x_i)} = \sum_{y \varepsilon Y} \sum_{x \varepsilon X} p(x_i, y_j) e^{-p^2(x_i)} \quad (5)$$

Similarly the proposed non-extensive entropy of Y is given by:

$$H(Y) = \sum_{y \varepsilon Y} p(y_j) e^{-p^2(y_j)} = \sum_{x \varepsilon X} \sum_{y \varepsilon Y} p(x_i, y_j) e^{-p^2(y_j)} \quad (6)$$

Then the *conditional entropy* of X given Y is defined as:

$$H(X|Y) = \sum_{x \varepsilon X} \sum_{y \varepsilon Y} p(x_i, y_j) e^{-p^2(x_i|y_j)} \quad (7)$$

Similarly the conditional entropy of Y given X is defined as:

$$H(Y|X) = \sum_{x \varepsilon X} \sum_{y \varepsilon Y} p(x_i, y_j) e^{-p^2(y_j|x_i)} \tag{8}$$

The *joint entropy* of X and Y is defined as:

$$H(X,Y) = \sum_{x \varepsilon X} \sum_{y \varepsilon Y} p(x_i, y_j) e^{-p^2(x_i, y_j)} \tag{9}$$

The *relative entropy* or *Kullback-Liebler information number* [20] is a measure of the relative distance between the two probability distributions and is defined by

$$D(X \| Y) = e^{-1} - \sum_{i=1}^{n} p(x_i) e^{-\frac{p(x_i)^2}{p(y_i)^2}} \tag{10}$$

The relative entropy is not necessarily symmetric i.e., $D(X \| Y) \neq D(Y \| X)$. Since $e^{-\frac{p(x_i)^2}{p(y_i)^2}} \geq 0$ and $0 \leq p(x_i) \leq 1$ in (10), the sum term, $\sum_{i=1}^{n} p(x_i) e^{-\frac{p(x_i)^2}{p(y_i)^2}}$ is non negative. Now the relation (10) is analyzed with respect to the two distributions $p(x_i), p(y_i)$.

If the two probability distributions $p(x_i), p(y_i)$ are the same then the relative entropy in (10) is at a minimum of $D(X \| Y) = 0$. This happens because $\sum_{i=1}^{n} p(x_i) = 1$ and $\sum_{i=1}^{n} p(x_i) e^{-\frac{p(x_i)^2}{p(y_i)^2}}$ evaluates to $e^{-1}$. On the other hand, $D(X \| Y)$ increases if the two distributions diverge having their values in the ranges [0,1] and [1,0] respectively, reaching a maximum of $e^{-1}$ when the sum $\sum_{i=1}^{n} p(x_i) e^{-\frac{p(x_i)^2}{p(y_i)^2}}$ evaluates to zero for the extreme limits. If p(y$_i$)=0 and p(x$_i$)=0 then the term $0 e^{-\frac{0}{0}}$ is evaluated as zero (quite similar to Shannon's assumption that 0log0=0).

## 3.2 Theorems regarding the properties of Non-Extensive entropy function

**Theorem 1** The proposed conditional entropy is bounded by the limits zero (lower bound) and the individual entropy (upper bound).

**Proof:** *a.)Lower Bound of Conditional entropy:*
Since $p(x_i, y_j) = p(x_i | y_j) p(y_j)$, the conditional entropy of X given Y in (7) can be written as

$$H(X|Y) = \sum_{x \varepsilon X} \sum_{y \varepsilon Y} p(x_i, y_j) e^{-p^2(x_i|y_j)} = \sum_{x \varepsilon X} \sum_{y \varepsilon Y} p(x_i, y_j) e^{-\frac{p^2(x_i, y_j)}{p^2(y_j)}} \tag{11}$$

Similarly,

$$H(Y|X) = \sum_{y \varepsilon Y} \sum_{x \varepsilon X} p(y_j, x_i) e^{-p^2(y_j|x_i)} = \sum_{y \varepsilon Y} \sum_{x \varepsilon X} p(x_i, y_j) e^{-\frac{p^2(x_i, y_j)}{p^2(x_i)}} \tag{12}$$

Since $0 \le p(x_i, y_j) \le 1$ and $e^{-\frac{p^2(x_i, y_j)}{p^2(y_j)}} > 0$ in (11) and $e^{-\frac{p^2(x_i, y_j)}{p^2(x_i)}} > 0$ in (12) therefore,

$$H(X|Y), H(Y|X) \ge 0 \qquad (13)$$

b.) *Upper Bound of Conditional entropy*

$$H(X) + H(X|Y) = \sum_{x \varepsilon X} p(x_i) e^{-p^2(x_i)} + \sum_{x \varepsilon X} \sum_{y \varepsilon Y} p(x_i, y_j) e^{-p^2(x_i|y_j)}$$

$$= \sum_{x \varepsilon X} \sum_{y \varepsilon Y} p(x_i, y_j) e^{-p^2(x_i)} + \sum_{x \varepsilon X} \sum_{y \varepsilon Y} p(x_i, y_j) e^{-\frac{p^2(x_i, y_j)}{p^2(y_j)}}$$

$$= \sum_{x \varepsilon X} \sum_{y \varepsilon Y} p(x_i, y_j) \left( e^{-p^2(x_i)} + e^{-\frac{p^2(x_i, y_j)}{p^2(y_j)}} \right) = \sum_{x \varepsilon X} \sum_{y \varepsilon Y} p(x_i, y_j) e^{-p^2(x_i)} \left( 1 + \frac{e^{-\frac{p^2(x_i, y_j)}{p^2(y_j)}}}{e^{-p^2(x_i)}} \right) \qquad (14)$$

$$= \sum_{x \varepsilon X} \sum_{y \varepsilon Y} p(x_i, y_j) e^{-p^2(x_i)} \left( 1 + e^{-\left( \frac{p^2(x_i, y_j) - p^2(x_i) p^2(y_j)}{p^2(y_j)} \right)} \right)$$

If X and Y are statistically independent, i.e. $p(x_i, y_j) = p(x_i) p(y_j)$, then (14) reduces to

$$H(X) + H(X|Y) = 2H(X)$$

or $\qquad H(X|Y) = H(X) \qquad (15)$

Thus the conditional entropy reduces to the individual entropy when the random variables are statistically independent thus conforming to the axioms of the classical probability theory. By a similar argument, for the statistically independent case,

$$H(Y) + H(Y|X) = 2H(Y)$$

or $\qquad H(Y|X) = H(Y)$

In the statistically dependent cases, $p(x_i|y_j) > p(x_i)$ therefore from (5) and (7) it can be deduced that:

$$H(X|Y) < H(X) \qquad (16)$$

i.e. the conditional entropy is less than the individual entropy for the dependent cases .
From (15) and (16), it is concluded that the upper bound of conditional entropy is the individual entropy itself and the conditional entropy exists between the limits $0 \le H(X|Y) \le H(X)$ and $0 \le H(Y|X) \le H(Y)$.

**Theorem 2** The proposed non-extensive entropy is non additive in nature
**Proof:**

$$H(X) + H(Y) = \sum_{x \varepsilon X} p(x_i) e^{-p^2(x_i)} + \sum_{y \varepsilon Y} p(y_j) e^{-p^2(y_j)}$$

$$\cong \sum_{x \varepsilon X}\sum_{y \varepsilon Y} p(x_i, y_j)\left( e^{-\sum_{y \varepsilon Y} p^2(x_i, y_j)} + e^{-\sum_{x \varepsilon X} p^2(x_i, y_j)} \right)$$

$$> \sum_{x \varepsilon X}\sum_{y \varepsilon Y} p(x_i, y_j) e^{-p^2(x_i, y_j)} \quad (17)$$

(as can be easily proved by Jensen's inequality)
From (9) and (17) it can be concluded that,

$$H(X) + H(Y) > H(X, Y) \quad (18)$$

Therefore the joint entropy of X and Y is always less than the sum of the individual entropies. It is observed that unlike Shannon entropy, the equality condition is not satisfied for the statistically independent case since $\left(e^{-p^2(x_i)} + e^{-p^2(y_j)}\right) \neq e^{-p^2(x_i) p^2(y_j)}$.

This non-additive nature of the proposed entropy is useful for representing non-additive information found in some random physical processes and makes it ideal for characterizing various random textures found in nature.

**Theorem 3** The proposed joint entropy is bounded by the (upper limit) sum of individual entropies and the (lower limit) conditional entropies $H(X|Y), H(Y|X)$ which are bounded as proved in Theorem 1.

**Proof:** The proof easily follows from (9) by substituting $p(x_i, y_j) = p(x_i | y_j) p(y_j)$ and $p(x_i, y_j) = p(y_j | x_i) p(x_i)$ separately to get the two results:

$$H(X,Y) = \sum_{x \varepsilon X}\sum_{y \varepsilon Y} p(x_i, y_j) e^{-p^2(x_i, y_j)} = \sum_{x \varepsilon X}\sum_{y \varepsilon Y} p(x_i, y_j) e^{-p^2(x_i | y_j) p^2(y_j)}$$

$$\geq \sum_{x \varepsilon X}\sum_{y \varepsilon Y} p(x_i, y_j) e^{-p^2(x_i | y_j)} \quad (19)$$

$$\geq H(X|Y)$$

$$H(X,Y) = \sum_{x \varepsilon X}\sum_{y \varepsilon Y} p(x_i, y_j) e^{-p^2(x_i, y_j)} = \sum_{x \varepsilon X}\sum_{y \varepsilon Y} p(x_i, y_j) e^{-p^2(y_j | x_i) p^2(x_i)}$$

$$\geq \sum_{x \varepsilon X}\sum_{y \varepsilon Y} p(x_i, y_j) e^{-p^2(y_j | x_i)} \quad (20)$$

$$\geq H(Y|X)$$

Therefore the lower limit of the joint entropy is constrained by the finite values of the conditional entropies (as per Theorem 1). The upper limit of the joint entropy is found from Theorem 2 to be the sum of the individual entropies. Therefore, $(H(X|Y), H(Y|X)) \leq H(X,Y) < (H(X) + H(Y))$.

**Remark:** Following from Theorems 1 and 3, the proposed non-extensive entropy of a system is bounded by two or more random variables with the finite upper and lower limits. The non-additive property of the non-extensive entropy proved in Theorem 2 has tremendous significance for the suitability the proposed entropy for texture classification.

## 4. APPLICATION OF NON ADDITIVITY OF THE PROPOSED ENTROPY FEATURE FOR TEXTURE CLASSIFICATION

The additive property of the logarithmic Shannon and Renyi entropies renders them unsuitable for representing textures containing repetitive or correlated patterns due to the non additive information content found in them. There are strong correlations between pixels of the same texture in terms of luminous levels and spatial orientations. In such situations measuring the average information content of a texture by the simple addition of individual contributions of pixels is not enough and some extension is required. This requirement of a non additive entropy for representing textures containing correlated patterns is met by the proposed non-extensive entropy function.

Recent developments in Statistical Mechanics [18] claim that the non-extensive entropy is more suitable than the extensive entropy for representing the amount of disorder in physical systems with long range interactions, long-time memories and fractal type structures. Such types of physical systems are termed as non-extensive systems. All natural textures are classified as non-extensive systems due to the long range pixel interactions (spatial as well as intensity), strong pixel correlations and the presence of fractal structures in repetitive texture patterns. Tsallis entropy generalizes Boltzman-Gibb-Shannon entropy to non-extensive systems [17]. However the choice of the Tsallis coefficient q remains an open field of study [23] since the non-extensive Tsallis entropy converges to the extensive Boltzman Gibb's entropy as $q \rightarrow 1$. Our work proposes a new form of non-extensive entropy which is better suited than Tsallis entropy for representing natural textures, since it does not converge to the extensive form under any condition. Another generalization of the Shannon entropy for non-extensive systems is the Pal and Pal entropy which has an exponential function with a linear exponent as the information gain. The non linearity of the proposed Gaussian information gain function is more than that of Pal and Pal, leading to a steeper fall in the graph as the probability of occurrence rises. This improves the selection of 'relevant' events whose probability of occurrence is low enough to be considered for the computation of amount of uncertainty in the system. Moreover the texture discriminating capability of the proposed entropy is more than that of Pal and Pal entropy due to the advantages of the non linear exponent over the linear exponent of the exponential. The pattern outlined using Pal and Pal entropy would be more unspecific or blurred than the proposed entropy. It is claimed that the proposed entropy is the most effective for the textures with long range pixel interactions and strong correlations which come under the category of non-extensive systems. It is the best for short range interactions since it allows for texture correlations even for the shortest displacement since even the most irregular textures have some amount of correlation even at the smallest level. Structured texture patterns exhibit regularity and orientation in long range zone while irregular patterns are dominant in the short range zone. In the case of weak pixel correlations in irregular textures the additivity property of Shannon and Renyi entropies works better than in the case of strong correlations, since the additivity property states that the individual entropies simply add up in the statistically independent case. However these two entropies perform poorly in the presence of strong pixel correlations where the statistical independence property fails.

The polar interaction map of a texture is one of the fundamental tools used to analyze the texture in the long range and short range zones. In [24] the authors have devised techniques for obtaining the feature based pixel interaction map (FBIM) computed from the extended gray level difference histogram (EGLDH) to measure the spatial dependence of a texture feature. The EGLDH shows the frequencies of the absolute gray level differences between pixels separated by a spacing vector. However the EGLDH suffers from the disadvantage that for irregular textures the technique fails. Since our work is aimed at introducing a new form of entropy feature and proving its efficiency, we focus on the FBIM using co-occurrence matrix instead of the EGLDH. The magnitude and angle of the spacing vector are arbitrary parameters which are set to the following ranges of values in our experiments: Magnitude of spacing vector $d=1$ to 31; Angle of spacing vector $\theta=0°$ to 315° with increments of 45° (angular resolution of co-occurrence matrix). This creates a 8x31 matrix $F(\theta,d)$ called the Polar interaction map with rows equal to value of $\theta$ and the columns equal to various d values (Fig.3). The intensity coded polar interaction map brings out the structure of the texture by plotting the feature values for different spacing vector. It is an ideal tool for comparison of texture features in terms of discernability of a specific texture pattern when the displacement approaches the period of the texture.

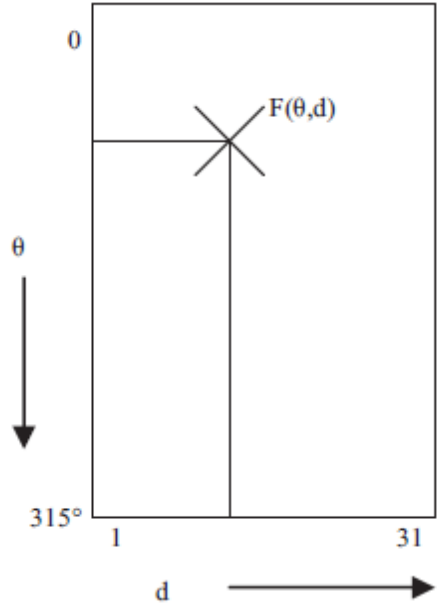

**Fig 3.** : *The structure of the Feature based polar interaction map(FBIM) where the feature value $F(\theta,d)$ is plotted for different values of the magnitude d and angle $\theta$ of the spacing vector.*

**5. METHODOLOGY FOR TEXTURE CLASSIFICATION**

Fifteen monochrome texture images of size 512x512 are selected from the Brodatz image database [22] and sixteen non-overlapping sub-images each of size 128x128 are obtained from each texture creating a database of 240 images out of which we randomly assign 120 as the training images and 120 as the test images. The gray level co-occurrence matrix [1] of the training set of texture images is calculated in four directions (namely, 0°, 45°, 90°, 135°). The entropy is computed from the gray level co-occurrence probabilities (GLCP) and averaged over the set of angles. The offset distance is set to 31 pixels for the 128x128 image since it is assumed that the texel unit or a significant portion of it is contained within one-eighth segment of the image. In the training phase, the entropy features extracted from the training images are used to train a SVM classifier [21] whose parameters are given in Table 1. These are selected for the best performance after several experiments on different types of kernel functions and the Pattern Search method is selected for optimizing the SVM parameters lambda, C and Epsilon. The trained SVM is applied on the entropy features computed from the test images for classification and the percentage of correct classification is computed. Cross-Validation is carried out by interchanging the training and the test datasets.

The performance of the proposed entropy is compared with that of other entropies like Shannon [15], Renyi [16], Tsallis [17] and Pal and Pal [19] entropies used for various image applications. While Shannon and Renyi entropies are logarithmic, Pal and Pal entropy is of an exponential nature and Tsallis entropy is a non-additive generalization of Shannon entropy. Pal and Pal entropy and Tsallis entropy were proposed especially for image segmentation [18,19] while Shannon entropy is popularly used in combination with other co-occurrence features for texture classification [1,2]. The four methods together constitute a good benchmark for evaluating the proposed entropy.

The performance of the proposed entropy feature is also evaluated against the Maximum response (MR4) filter method proposed by Varma and Zisserman in [28] which is one of the benchmark methods for texture classification. The filter size is 49x49 for six orientations at a fixed scale for two anisotropic filters(edge and bar filters) and two additional isotropic filters(Gaussian and LoG filters), and the maximum response of the filter over all orientations for each of the anisotropic filters together with the direct responses of the two isotropic filters is selected as the feature set for classification. The pair-wise pixel interactions for each texture are compiled as a polar feature based interaction map (FBIM) for different angles and magnitudes of the spacing vector. In this work the values of the five computed entropies are converted into a 8x31 matrix as explained in Section 4 above and this matrix is intensity coded to display as an image. The pixel correlation for each texture is computed from the gray level co-occurrence probabilities (GLCP) as per the formula:

$$COR = \sum_{i \varepsilon R} \sum_{j \varepsilon R} (i - \mu_x)(j - \mu_y) f_{ij} / (\sigma_x . \sigma_y) \tag{21}$$

where, R is the number of bins in the co-occurrence histogram (set equal to 256 to match the total number of gray intensity levels in an image), $f_{ij}$ is the GLCP computed between the gray levels $i$ and $j$ for a given spacing vector and $\mu_x, \mu_y$ are the horizontal and vertical mean of the co-occurrence matrix respectively and $\sigma_x, \sigma_y$ are the horizontal and vertical standard deviations of the co-occurrence matrix respectively. The 8x31 polar interaction map based on the correlation feature is computed as described in Section 4 and is plotted as an intensity image. The result is compared with the entropy based polar interaction maps to find the best match between texture correlation and texture entropy.

## 6. EXPERIMENTAL RESULTS AND DISCUSSIONS

The experiments are conducted on fifteen 512x512 monochrome textures from the Brodatz texture [22] database (Fig. 4) which includes both grainy as well as low frequency type of textures.

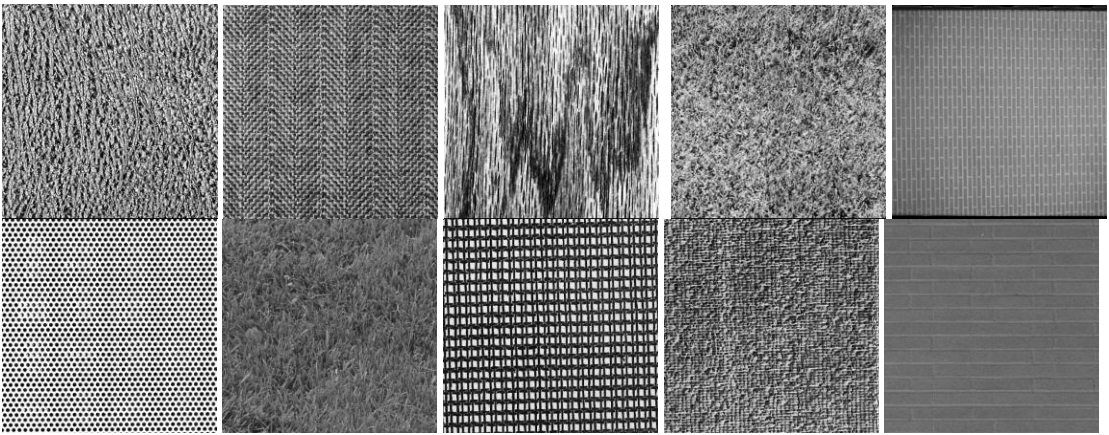

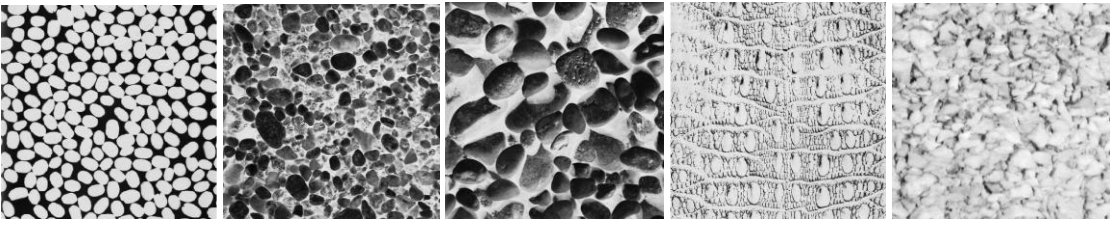

**Fig 4.** : *Brodatz Texture images. From left to right and top to bottom: Pressed Calf Leather(D24), Herringbone Weave(D16), Wood Grain(D68), Grass (D9),Brick Wall (1.4.01), Hexagonal hole array(1.5.02), Grass (1.5.07), French Canvas(D20), Raffia (D84), Brick wall(1.5.06),D75,D23,D31,D10,1.5.05*

The gray level co-occurrence matrix is computed for all the 16 sub-images obtained from each of the fifteen textures as explained in Section 5. The entropy feature is computed from the gray level co-occurrence probabilities and 120x1 training feature vector is used to train the SVM multi-class classifier. The 120x1 test feature vector is then applied as input to the trained SVM classifier for classification purpose and the results are listed in Table 2. It is observed that the proposed non-extensive entropy yields the highest average correct classification rate of 98.33% as compared to that of the Shannon (96.43%), Renyi(96.02%), Tsallis (95.63%) and Pal and Pal (95.10%) entropies for the validation data indicating the effectiveness of the proposed method for the texture classification problem. The cross-validation also yields the highest classification accuracy rate for the proposed entropy as shown in Table 2. The mean value of the proposed entropy function for each texture (averaged over 16 sub-images) and its standard deviation are shown in Table 3. It is noted that the regular textures like *D20* and *1.5.02* yield a low value for the proposed entropy indicating their repetitive texture patterns. On the other hand, textures *D23, D24, 1.5.07* and *D9* yield a higher value for the non-extensive entropy indicating random structures. Table 3 also shows that the standard deviation values of the proposed entropy are sufficiently small resulting in high classification accuracy. Table 4 compares the performance of the proposed entropy with the MR4 filter technique for texture classification and the results are shown for both Validation and Cross-Validation data. It is observed that the proposed entropy outperforms the MR4 technique and gives a consistently high classification rate for all textures while MR4 technique performs poorly for random textures like *D9, D31* and *D23*. The proposed entropy is sensitive to the scale of the texture since beyond a certain limit certain textures start resembling each other. However to some extent rotation invariance is achieved since the proposed feature extraction scheme involves averaging of the feature over four angles. This is verified from the fact that the proposed entropy based technique outperforms the rotationally invariant MR4 filter technique which computes the maximum filter response over six orientations at a fixed scale. Representation of a texture by a single entropy feature is also a primary advantage since several recent works [30,31] are aimed at optimizing the feature dimension for fast image processing.

The polar feature based interaction maps (FBIMs) of the fifteen textures based on the correlation and the five entropies are plotted. The FBIMs are intensity coded to display as images shown in Fig. 5. The presence of dark blobs in the entropy FBIMs indicates regions of maximum interaction or correlation of pixels. The displacement or spacing at which the blobs occur corresponds to the approximate period of the texture and determines whether the correlations are long range or short range. In the case of textures with irregular or random patterns (*D68, D84*), the short range pixel interactions are dominant (Fig. 5 (iv, ix)) whereas for regular, structured patterns (*1.4.01, 1.5.06*), the interaction is observed at long range distances (Fig. 5(v, x)). Low frequency textures like *D31, 1.5.06* containing 'pebble' type structures are also correctly classified with high accuracy together with the grainy ones like *D24* and *D9*. There is a third category of textures called hierarchical textures (*D16, 1.5.02, D20,D10*) with a directional pattern at both long and short ranges(Fig. 5(ii, vi, viii, xiv)). The proposed entropy is able to distinguish all the directional components of the hierarchical textures. Additive entropies perform poorly in the case of this mixed type of texture structure. The non additive Tsallis and Pal and Pal entropies perform better. However while the Tsallis entropy interaction maps have too many variations, the Pal and Pal entropy is blurred since not much difference is found amongst the high (and low) entropy values. Shannon and Renyi entropy perform poorly in the case of long range strong pixel correlations as seen in Fig. 5(v) and are relatively better for short range strong pixel correlations (Fig. 5(i)). Renyi entropy outperforms Shannon entropy in all cases. The proposed entropy performs well in both the short and long range zones predicting the texture orientation accurately as seen from the FBIMs and the spacing at which maximum pixel correlation exists is observed from the interaction maps. Thus the proposed entropy can identify a particular texture with a high degree of accuracy.

The pixel correlation feature based interaction map is compared with the entropy based interaction maps to search for a good match. The white blobs in the correlation interaction map indicating high pixel correlations should ideally correspond to the dark blobs in the entropy interaction map indicating areas of minimum entropy or maximum regularity. As can be seen from Fig. 5 the proposed non-extensive entropy offers the best visual match with the correlation values as compared with all other entropies. The proposed entropy is a better indicator of high correlation areas due to the larger blobs formed (Fig. 5(vi)) and there is a clear distinction of the blobs from the background due to the non linearity of the non-extensive entropy (Fig. 5(v)). To compare the proposed entropy values in Table 3 with the correlation, we arrange the fifteen textures in the increasing order of entropy in Table 5 starting with the most regular (*1.5.02*) texture and ending with the most irregular (*D23*) texture. The texture period or displacement *d* at which the strong pixel correlations or white blobs are observed in the correlation FBIMs in Fig 5 and the value of the correlation at the location of the white blobs are compiled for each texture in Table 5 . The range of pixel interaction (long, short, long & short) based on the location of white blobs and the corresponding type of texture (regular, irregular, hierarchical) is given next. From Table 5, we arrive at the following conclusions:

1. The minimum value of proposed entropy is obtained for the set of Hierarchical textures with both long range and short range correlations, followed by the set of regular textures with long range correlations and the maximum values for the set of irregular or short range zone textures.

2. A high correlation measure for a texture is indicative of a repetitive gray level pattern in the texture contributing to regularity in structure. However, the angle and displacement at which maximum correlation is obtained also plays a role in predicting the randomness or entropy of an image since a highly correlated but very closely spaced texture pattern may make the texture appear grainy and give a high value of entropy. The proposed entropy provides a good match with the correlation of a texture based on both the gray level values and spatial arrangement of the texture pattern as shown in Table 5. For texture D23 no distinct blob is present as observed from Table 5 and Fig. 5 (xii) and thus it is the least correlated texture having the maximum value of proposed entropy. Within a particular set, the proposed entropy is inversely proportional to the spatial and gray level correlation therefore providing an accurate measure of the overall pixel correlation at the texture period.

We summarize from the above observations that the proposed entropy clearly demarcates among the hierarchical, regular and irregular texture groups and also within each group where the textures are graded depending on the strength of the combination of gray-level and spatial correlation at the texture period with the strongest correlation assigned the minimum entropy and vice versa.

It may be noted that the work of this paper is directed towards establishing a new form of non-extensive entropy and to investigate its superior ability to discriminate between various textures of different spatial and intensity correlations as compared to the existing forms of entropy. It is concluded from the experimental results that the proposed entropy is capable of representing high pixel correlations most accurately and hence gives better classification results.

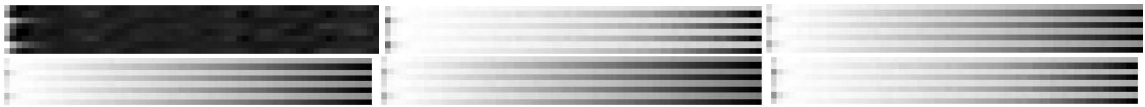
(i)

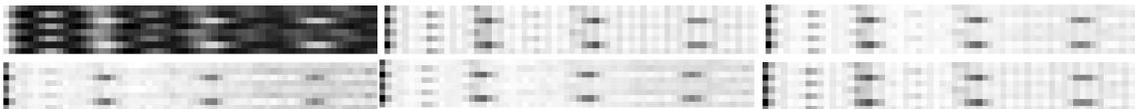
(ii)

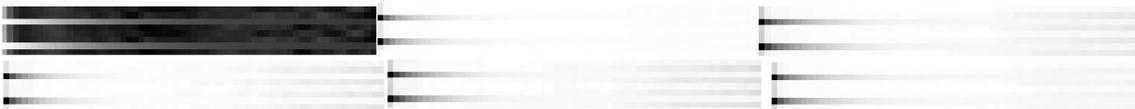
(iii)

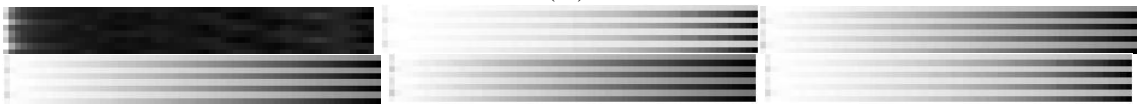
(iv)

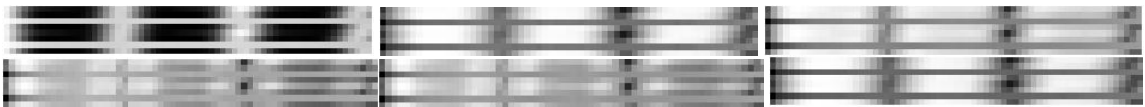
(v)

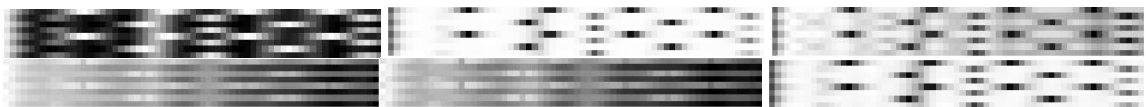
(vi)

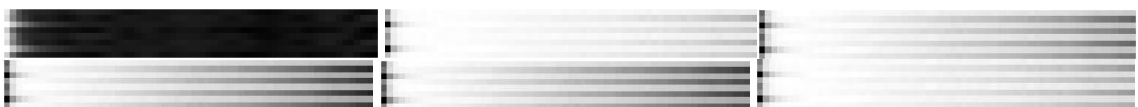
(vii)

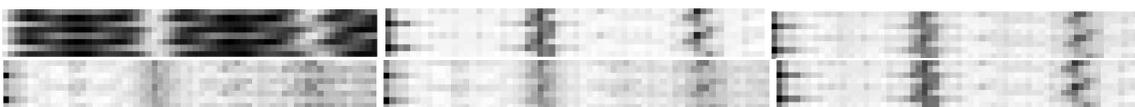
(viii)

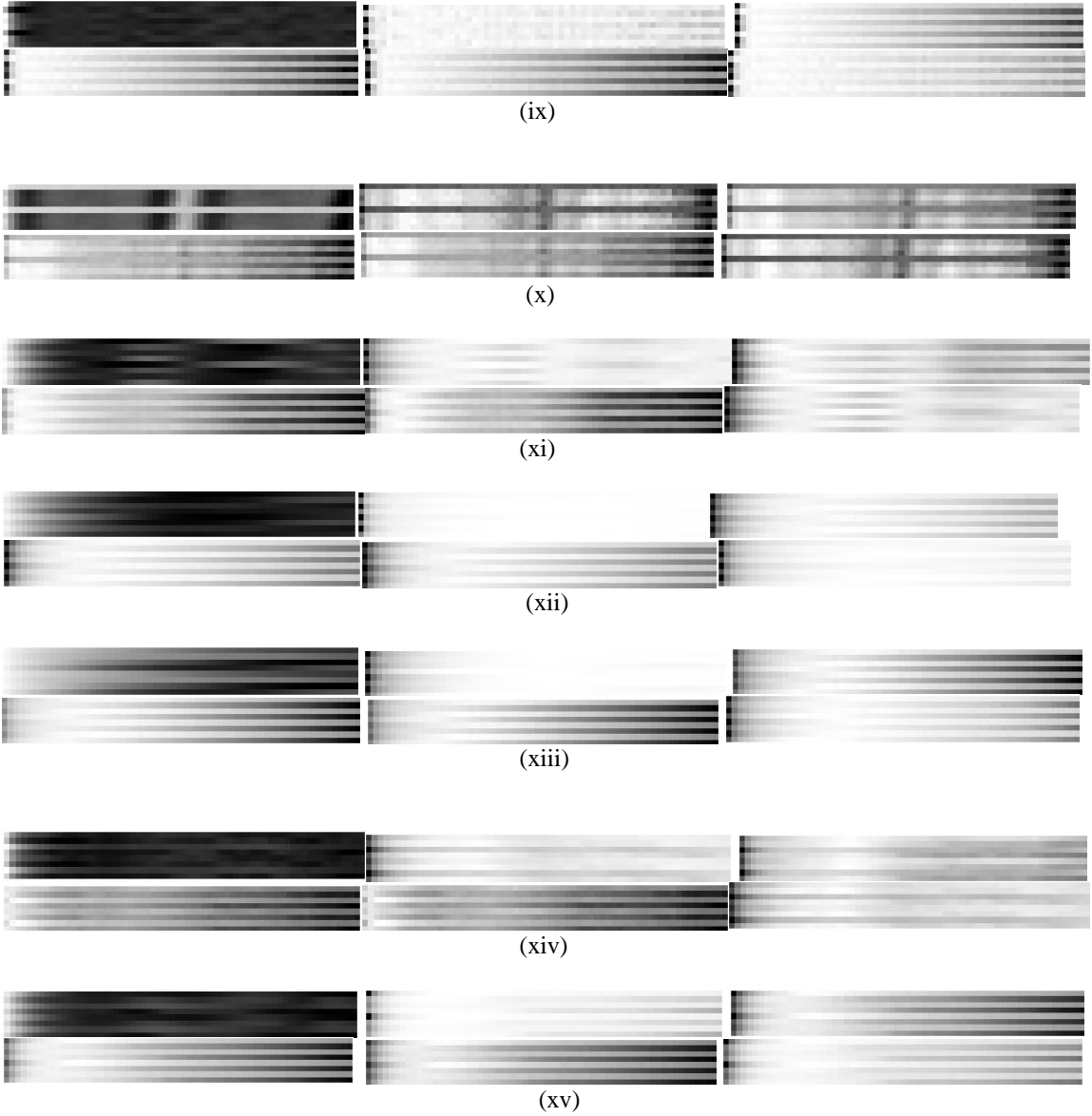

**Fig. 5**: *The feature based polar interaction maps (FBIM) of the fifteen textures in Fig 4 in the order (i) to (xv) with the following order of features for each texture: Correlation, Proposed Non-extensive Entropy, Shannon's Entropy, Renyi's Entropy, Tsallis's Entropy and Pal's Entropy.*

**Table 1:** Multi-class SVM training parameters

| | |
|---|---|
| Number of Training inputs | 120 |
| Number of Test inputs | 120 |
| Type of SVM model | C-SVC classification model with a Categorical target variable |
| Method for finding separating hyperplane | Quadratic Programming (QP) |
| Kernel Function | Gaussian Radial Basis Function |
| Method for refining parameters | Pattern Search (PS) [10 intervals, $10^{-8}$ tolerance] |
| Conditioning parameter for the QP method | lambda=447 |
| Tolerance to mis-classification errors | C = 3 |
| Epsilon parameter | 0.001 |
| Type of Validation used | Cross Validation of training and test datasets |

**Table 2:** The percentage of correct classification after Validation (V) and Cross-Validation (CV) for proposed entropy and Shannon, Renyi, Tsallis and Pal &Pal entropies

| Texture | Proposed Entropy | | Shannon Entropy | | Renyi Entropy | | Tsallis Entropy | | Pal & Pal Entropy | |
|---|---|---|---|---|---|---|---|---|---|---|
| | V | CV | V | CV | V | CV | V | CV | V | CV |
| D24 | 100% | 100% | 98.33% | 98.33% | 96.17% | 97.5% | 97.5% | 93.33% | 93.3% | 93.3% |
| D16 | 98.33% | 100% | 100% | 100% | 94.16% | 95.83% | 95.83% | 93.3% | 96.67% | 97.50% |
| D68 | 99.17% | 98.33% | 99.16% | 96.67% | 100% | 100% | 99.16% | 98.33% | 96.67% | 94.17% |
| D9 | 98.33% | 99.17% | 100% | 100% | 99.17% | 100% | 97.5% | 98.33% | 93.3% | 93.3% |
| 1.4.01 | 98.33% | 95.83% | 99.16% | 99.16% | 95% | 93.3% | 94.16% | 95% | 99.17% | 95.83% |
| 1.5.02 | 100% | 99.17% | 92.50% | 93.3% | 98.33% | 96.67% | 95% | 94.16% | 89.17% | 91.67% |
| 1.5.07 | 98.33% | 98.33% | 97.50% | 91.67% | 96.67% | 89.17% | 94.16% | 93.33% | 93.3% | 93.3% |
| D20 | 99.17% | 96.67% | 100% | 98.33% | 92.5% | 92.5% | 95.83% | 94.17% | 91.67% | 91.67% |
| D84 | 98.33% | 99.17% | 92.5% | 93.3% | 93.3% | 93.3% | 93.3% | 93.3% | 95% | 93.3% |
| 1.5.06 | 98.33% | 95.83% | 95% | 95% | 97,5% | 97.5% | 99.16% | 99.16% | 100% | 100% |
| D75 | 96.67% | 99.17% | 90% | 90% | 95.83% | 95.83% | 93.3% | 93.3% | 95% | 92.5% |
| D23 | 97.5% | 100% | 99.16% | 100% | 99.17% | 100% | 95.83% | 97.5% | 99.16% | 100% |
| D31 | 98.33% | 96.67% | 94.17% | 96.67% | 92.5% | 95.83% | 94.16% | 93.3% | 96.67% | 95.83% |
| D10 | 97.50% | 95.83% | 96.6% | 98.33% | 97.5% | 95.83% | 96.67% | 97.5% | 94.17% | 93.3% |
| 1.5.05 | 96.67% | 96.67% | 92.50% | 87.50% | 92.5% | 89.17% | 93.3% | 93.3% | 93.33% | 93.33% |
| **Average** | **98.33%** | **98.05%** | **96.43%** | **95.88%** | **96.02%** | **95.49%** | **95.63%** | **95.15%** | **95.10%** | **94.59%** |

**Table 3**: The proposed entropy values and their standard deviations for different textures

| Texture | Mean Value of Proposed entropy | Standard deviation of Proposed entropy |
|---|---|---|
| D24 | 0.999999949476079 | 2.01842612529080e-09 |
| D16 | 0.999856024675990 | 1.72554047955097e-05 |
| D68 | 0.999987974569801 | 4.70978634383390e-06 |
| D9 | 0.999999971216022 | 1.74686601411381e-09 |
| 1.4.01 | 0.999996121007935 | 1.28480412097562e-06 |
| 1.5.02 | 0.996587666291163 | 0.00146760985742619 |
| 1.5.07 | 0.999999903834812 | 1.57763515894054e-08 |
| D20 | 0.999241367860436 | 0.000130549018665018 |

| Texture | | | | |
|---|---|---|---|---|
| D84 | 0.999999725867491 | | 4.51429032310604e-08 | |
| 1.5.06 | 0.999995337137991 | | 4.65832565428011e-07 | |
| D75 | 0.999947742424632 | | 3.35226056415152e-05 | |
| D23 | 0.999999982516793 | | 1.68234743446299e-08 | |
| D31 | 0.999998144438016 | | 1.73561560549102e-06 | |
| D10 | 0.999594302556659 | | 0.000256120511314726 | |
| 1.5.05 | 0.999999832216796 | | 9.38931745093621e-08 | |

**Table 4**: The percentage of correct classification after Validation (V) with a single training sample and fifteen test samples for each texture and Cross-Validation (CV) for proposed entropy and MR4 filter based method.

| Texture | Proposed Entropy | | MR4 Filter | |
|---|---|---|---|---|
| | V | CV | V | CV |
| D24 | 99.56% | 100% | 99.56% | 100% |
| D16 | 98.22% | 100% | 96% | 100% |
| D68 | 98.67% | 100% | 96.44% | 93.33% |
| D9 | 99.11% | 100% | 93.78% | 100% |
| 1.4.01 | 95.56% | 100% | 100% | 100% |
| 1.5.02 | 98.67% | 100% | 100% | 100% |
| 1.5.07 | 99.11% | 100% | 99.56% | 100% |
| D20 | 96.89% | 100% | 99.56% | 100% |
| D84 | 97.33% | 100% | 93.33% | 93.33% |
| 1.5.06 | 96.44% | 100% | 100% | 100% |
| D75 | 98.22% | 100% | 98.22% | 100% |
| D23 | 98.67% | 100% | 95.56% | 100% |
| D31 | 96.89% | 100% | 95.11% | 100% |
| D10 | 95.11% | 100% | 95.11% | 86.67% |
| 1.5.05 | 98.22% | 100% | 96.89% | 100% |
| **Average** | **97.78%** | **100%** | **97.27%** | **91.55%** |

**Table 5:** The relation between the proposed entropy and the texture correlation

| Texture (in increasing order of proposed entropy) | Spacing of white blobs in correlation FBIM (Fig. 5) | Gray-level Correlation Value at Location of white blobs | Range of Interaction (from d in first column) | Type of Texture (based on range of Interaction) |
|---|---|---|---|---|
| 1.5.02 | d=14, θ=0° <br> d=24, θ=90° <br> d=28, θ=0° <br> d=35, ,θ=45° | 0.9861 <br> 0.9775 <br> 0.9589 <br> 0.8842 | Long & short | Hierarchical |
| D20 | d=1, θ=90° <br> d=26, θ=0° | 0.9085 <br> 0.8241 | Long & short | Hierarchical |
| D10 | d=1, θ=90° <br> d=15, θ=0° <br> d=31, θ=90° | 0.8500 <br> 0.0119 <br> 0.1070 | Long & short | Hierarchical |
| D16 | d=8, θ=45° <br> d=16, θ=0° | 0.4541 <br> 0.4203 | Long & short | Hierarchical |
| D75 | d=30,θ=135° <br> d=45, θ=0° | 0.1481 <br> 0.2816 | Long & short | Hierarchical |
| D68 | d=1, θ=90° <br> d=10 θ=0° <br> d=10 θ=45° | 0.9626 <br> 0.0977 <br> 0.0735 | Long & short | Hierarchical |
| D31 | d=19 θ=0° <br> d=51,θ=135 | 0.4143 <br> 0.0382 | Long & short | Hierarchical |
| 1.5.06 | d=33, θ=0° | 0.4062 | Long | Regular |
| 1.4.01 | d=25, θ=90° | 0.7445 | Long | Regular |
| D84 | d=11, θ=0° | 0.0262 | Long | Regular |
| 1.5.05 | d=1, θ=90° | 0.9540 | Short | Irregular |
| 1.5.07 | d=1, θ=90° | 0.8873 | Short | Irregular |
| D24 | d=1, θ=90° | 0.7706 | Short | Irregular |
| D9 | d=1, θ=0° <br> d=1, θ=90° | 0.7289 <br> 0.7278 | Short | Irregular |
| D23 | d=1, θ=0° <br> d=1, θ=45° <br> d=1, θ=90° <br> d=1, θ=135° | 0.9798 <br> 0.9604 <br> 0.9763 <br> 0.9586 | Short | Irregular |

## 7. CONCLUSIONS

A new non-extensive entropy measure is devised in this paper based on a Gaussian information gain function, and its various properties are derived. It is claimed that the proposed entropy function is ideal for texture identification and classification. The non additivity property of the proposed entropy function makes it ideal for the representation of a non-extensive system involving correlated random variables. The effectiveness of our approach is vindicated by applying the new entropy function for the classification of natural textures from the Brodatz texture database since all textures come under non-extensive systems with some degree of correlation within them. The entropy function is computed from the gray level co-occurrence probabilities. The non-extensive entropy function classifies textures with high degree of accuracy as compared to other well known forms of entropy. From the polar feature based interaction maps (FBIM) it is verified that the proposed entropy performs well under both long range and short range pixel correlations and hence is ideal for representing both regular and irregular textures.